\documentclass[letterpaper, 10 pt, conference]{ieeeconf}
\usepackage{amsmath,graphicx}
\usepackage[aboveskip=0pt,belowskip=-4pt]{caption}
\usepackage{url}
\usepackage{hyperref}
\usepackage[table]{xcolor}
\usepackage{color}
\usepackage[table]{xcolor}
\usepackage{booktabs}
\newcommand{\mypar}[1]{\vspace{0.15em}
\textbf{#1}~}

\IEEEoverridecommandlockouts

\title{Deep Modeling and Interpretation for Bladder Cancer Classification}

\author{
Ahmad Chaddad$^{1,2,*}$\thanks{This research was funded by the National Natural Science Foundation of China (Grant No. 82260360), and the Guangxi Science and Technology Base and Talent Project (Grant Nos. 2022AC18004 and 2022AC21040).}, 
Yihang Wu$^{1}$, 
Xianrui Chen$^{1}$ \\
$^1$Laboratory for AIPM, School of Artificial Intelligence, Guilin University of Electronic Technology, China \\
$^2$Laboratory for Imagery, Vision and Artificial Intelligence, École de Technologie Supérieure, Canada \\
All authors contributed equally to this work, *Corresponding author: \texttt{ahmad8chaddad@gmail.com}
}

\begin{document}

\maketitle
\begin{abstract}
Deep models based on vision transformer (ViT) and convolutional neural network (CNN) have demonstrated remarkable performance on natural datasets. However, these models may not be similar in medical imaging, where abnormal regions cover only a small portion of the image. This challenge motivates this study to investigate the latest deep models for bladder cancer classification tasks. We propose the following to evaluate these deep models: 1) standard classification using 13 models (four CNNs and eight transormer-based models), 2) calibration analysis to examine if these models are well calibrated for bladder cancer classification, and 3) we use GradCAM++ to evaluate the interpretability of these models for clinical diagnosis. We simulate $\sim 300$ experiments on a publicly multicenter bladder cancer dataset, and the experimental results demonstrate that the ConvNext series indicate limited generalization ability to classify bladder cancer images (e.g., $\sim 60\%$ accuracy). In addition, ViTs show better calibration effects compared to ConvNext and swin transformer series. We also involve test time augmentation to improve the models interpretability. Finally, no model provides a one-size-fits-all solution for a feasible interpretable model. ConvNext series are suitable for in-distribution samples, while ViT and its variants are suitable for interpreting out-of-distribution samples.  The codes are available at \url{https://github.com/AIPMLab/SkinCancerSimulation}.

\end{abstract}
\section{Introduction}
Cancers such as bladder cancer are major health concerns around the world, affecting people quality of life and general health status \cite{wang2024sscd}. In recent years, artificial intelligence (AI) has made rapid advancements in image processing, with Convolutional Neural Networks (CNNs) demonstrating the most significant impact \cite{chaddad2023federated}. Considering the limitations of traditional cancer diagnostic techniques, AI has the potential to enhance the accuracy of cancer classification \cite{wu2024facmic,10902405}. For instance, in \cite{jiao2024prediction}, ResNet50 was used as a feature extractor to convert each small patch of the bladder pathology image into a 1024-dimensional feature vector to provide data for subsequent prediction. In \cite{rippa2024classification}, they proposed to use ConvNext models to predict types of prostate cancer using MRI data. Their method yields an average precision score of 0.4583 and AUC ROC score of 0.6214 on a private dataset. However, given the variety of CNN architectures available, it is still unclear which is the most effective for bladder cancer, where radiologists have limited high-quality annotation \cite{cao2024multicenter}. The accurate diagnosis of bladder cancer is crucial for early treatment and improving patient survival rates and quality of life. Furthermore, unlike common natural datasets, high accuracy does not always equal a high diagnosis rate for rare diseases, especially when the data set is highly imbalanced. Even if the classifier model performs effectively on natural images, it does not guarantee similar performance on medical images.

Despite these studies on cancer classification, there are four challenges: 1) no comprehensive empirical study demonstrates the importance of CNN and ViT models with different optimizers using bladder cancer classification datasets, especially for multi-center based bladder cancer dataset that has considerable feature shifts. 2) Studies related to calibrating bladder cancer classifiers are limited, and it is still unclear which model can provide feasible calibration metrics for bladder cancer classification. 3) The interpretability of CNN and ViT for bladder cancer classification has not been investigated, preventing the real-world deployment of these models.

Motivated by these challenges, we aim to explore recent deep-network models to classify images of bladder cancer. Specifically, we considered 13 deep models with five optimizers to evaluate their potential for bladder cancer classification using classification metrics and execution time. In addition, we introduce the expected calibration error (ECE) and the reliability plot to investigate the calibration ability of these models. Finally, we incorporate GradCAM++ to visualize the key features obtained by these models to improve the transparency of the model predictions in both in-distribution and out-of-distribution settings. We also introduce test-time augmentation (TTA) to enhance the visual explanations of these models. Identifying the model that delivers the highest performance will assist physicians in making accurate and rapid diagnoses. The contributions of this paper can be summarized as follows.
\begin{enumerate}
    \item We propose 13 widely used deep CNNs and eight ViT based models, utilizing a multi-center public bladder cancer dataset and five different optimizers to predict cancer types. 
    \item we perform calibration analysis to validate the 13 models and employ GradCAM++ for interpreting the CNNs and ViTs. To further enhance interpretability, we apply Test Time Augmentation (TTA) across the 13 deep models.
\end{enumerate}

\begin{figure*}[!ht]
    \centering
    \includegraphics[width=\linewidth]{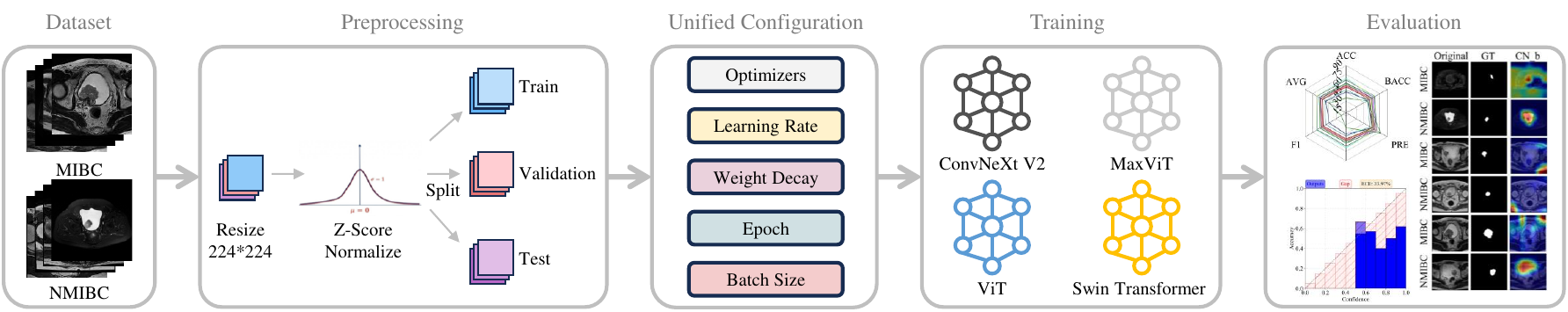}
    \caption{Flowchart illustrates the comparative analysis workflow for 13 models. The pipeline begins with images collection, followed by data preprocessing and model training. Finally, computation of performance metrics for each model to determine the most effective model in different situations.}
    \label{fig:flowchart}
\end{figure*}

\section{Related work}

\subsection{Deep CNNs based approaches}
In recent years, deep CNNs have become powerful tools for the classification of cancer images, taking advantage of their ability to automatically learn complex features from images \cite{ bashkami2024review}. For example, in \cite{zou2025prediction}, they introduce a self-distillation model of multiview fusion based on T2-weighted MRI images to predict types of bladder cancer. The experimental results suggest that their model provides an area under the curve (AUC) of 0.927 and accuracy of 0.880, respectively. Furthermore, in \cite{liu2021multiscale}, a multiscale ensemble of CNNs was proposed with the integration of attention mechanisms for the classification of cancer images, using pre-trained models as feature extraction modules. Recently, for bladder cancer images, in \cite{yang2021application}, they proposed a CNN model to classify muscle-invasive and non-muscle-invasive bladder cancer using contrast-enhanced CT images. Furthermore, in \cite{li2023predicting}, they propose the use of ResNet50 to predict the status of muscle invasive bladder cancer based on T2-weighted images. The experimental results indicate that it provided an area under the curve (AUC) of 0.93. Meanwhile, in \cite{shalata2024precise}, they use a multiscale pyramidal CNN architecture based on ShuffleNet to objectively classify non-muscle invasive bladder cancer by analyzing digital pathology images. In \cite{peng2025stacking}, they propose to use radiomic features and deep features with a set method to segment and classify images of bladder cancer. Moreover, the study in \cite{eminaga2023efficient} illustrates that lightweight models, such as MobileNet and PlexusNet, deliver lesion detection accuracy comparable to that of more complex architectures such as ConvNeXt and SwinTransformer, achieving 100\% sensitivity at block/ROI levels in real-time bladder cancer screening.

\subsection{Transformer based approaches}
Similar to CNNs, transformer-based models have been used for cancer diagnosis. For example, in \cite{guergueb2022skin}, they explore the detection of melanoma cancer using ensemble learning and Swin Transformer V2 among other models, achieving high accuracy and robustness. In \cite{khan2023skinvit}, they explore MaxViT for cancer classification, achieving 84.18\% accuracy on the ISIC2019 subset. Furthermore, in \cite{zhang2025texture}, the transformer model was selected for its ability to capture spatial dependencies between image regions, illustrating its promise to improve classification accuracy and generalization in bladder cancer classification tasks. For bladder cancer images, in \cite{khedr2023classification}, a dataset of histopathological images of bladder tissue samples was used to train ViT-based models for bladder cancer classification. The experimental results indicate that ViT-B/32 provided an accuracy of 99.49\%. In \cite{kurata2024development}, they focused on the diagnosis of muscle invasive bladder cancer using a multicenter retrospective MRI data set. In \cite{el2025accurate}, they propose an ensemble approach by combining CNNs with a transformer for an accurate diagnosis of bladder cancer. The experimental results suggest that it provides the highest accuracy and the lowest execution time.

Unlike the previous studies, this study evaluates 13 deep models with five optimizers on one multi-center bladder cancer dataset, covering various situations such as traditional deep learning, classifier calibration, execution time, and interpretability analysis, aiming to provide a comprehensive understanding of these models for bladder cancer classification.  

\section{Methods}\label{S:3}

\mypar{Pipeline} Figure \ref{fig:flowchart} illustrates a flow chart comprising four main steps: 1) datasets and image pre-processing, 2) unified configuration, 3) training, and 4) evaluation.

\begin{table}[t] \scriptsize
    \centering
    \renewcommand{\arraystretch}{0.5} 
    \caption{Summary of hyper-parameter settings in optimizer.}
  \rowcolors{2}{gray!10}{white}
    \setlength{\tabcolsep}{18pt}
    \begin{tabular}{cccc}
    \toprule
         &LR&WD&Betas \\
    \midrule
    SGD&0.01&0.0004&- \\
    Adam&0.001&0.02&(0.9,0.98)\\ 
    AdamW&0.001&0.02&(0.9,0.98)\\ 
    Adagrad& 0.001&0.0005&- \\
    Adadelta&0.001&0.0005&- \\
    \bottomrule
    \end{tabular}
    \label{tab:optimizers}
\end{table}

\begin{figure}
    \centering
    \includegraphics[width=0.97\linewidth]{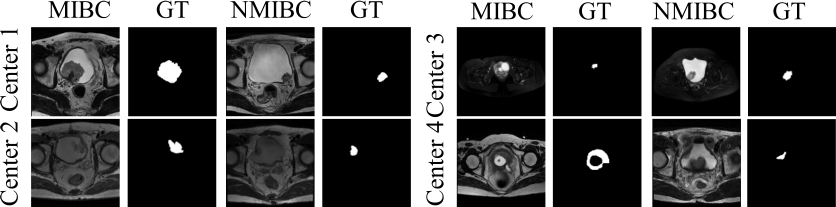}
    \caption{Example of MIBC and NMIBC samples in each center.}
    \label{fig:Datasets}
\end{figure}

\begin{figure*}[!ht]
    \centering
    \includegraphics[width=0.98\linewidth]{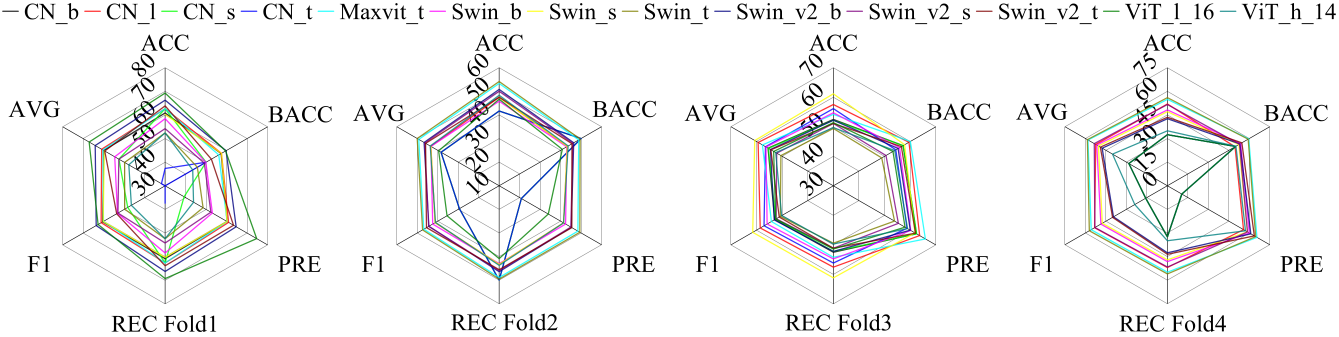}\\
    \vspace{1pt}\includegraphics[width=0.98\linewidth]{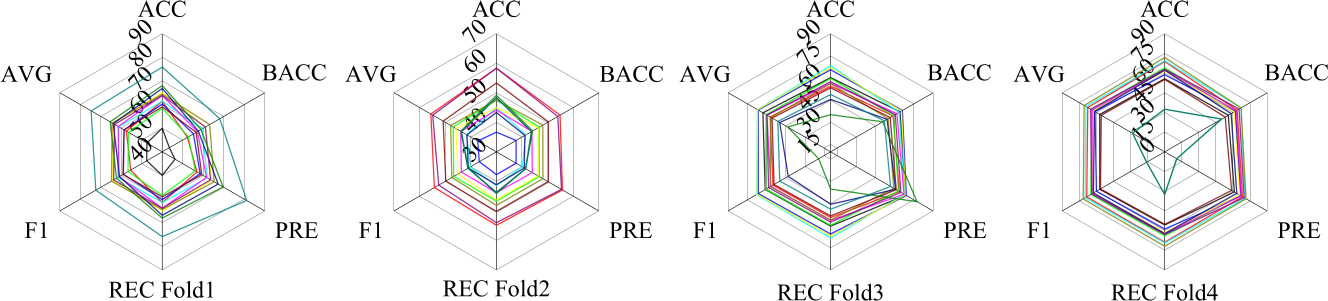}\\
    \vspace{1pt}\includegraphics[width=0.98\linewidth]{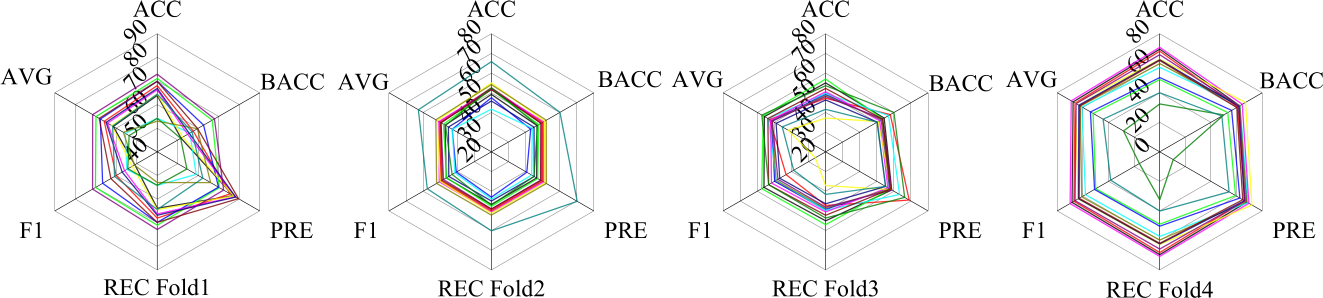}\\
    \vspace{1pt}\includegraphics[width=0.98\linewidth]{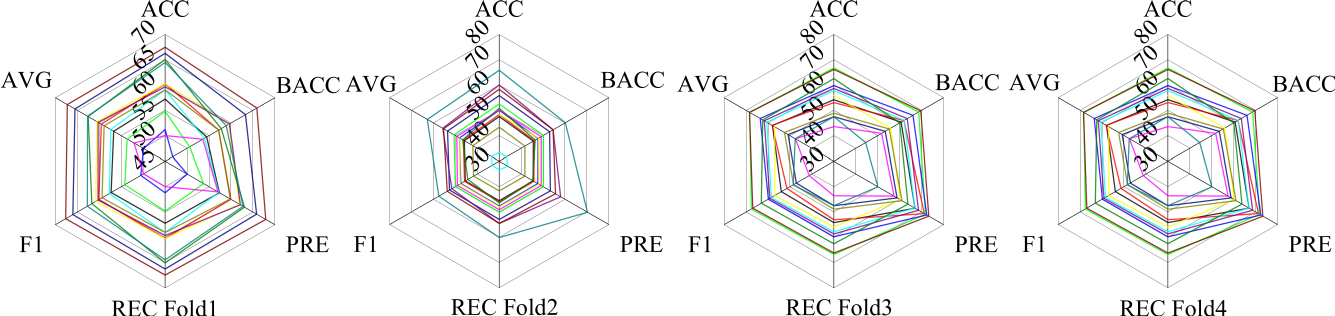}\\
    \vspace{1pt}\includegraphics[width=0.98\linewidth]{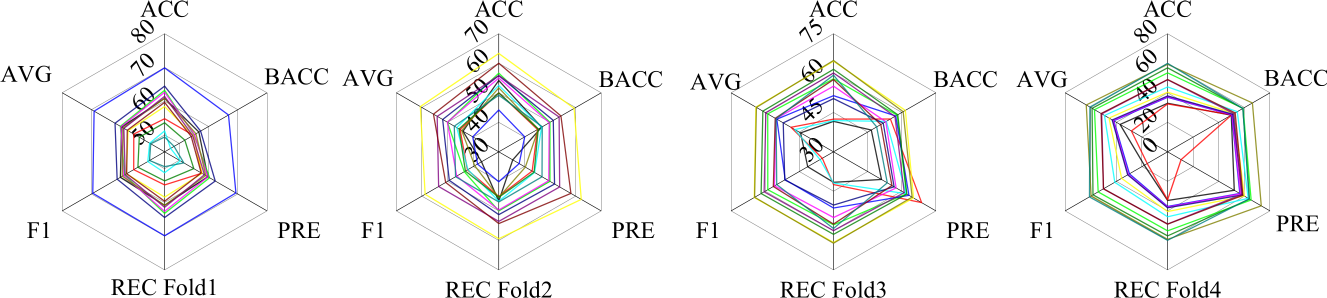}
    \caption{Spider plot of test metrics using SGD (\textbf{First row}), Adam (\textbf{Second row}), AdamW (\textbf{Third row}), Adagrad (\textbf{Fouth row}) and Adadelta (\textbf{Fifth row}) on Bladder dataset with 13 deep models.}
    \label{fig:Adagrad_Bladder_Spider}
\end{figure*}

\mypar{Image preprocessing.} For data input, this study uses z-score normalization to normalize the pixel values for all images. Additionally, we adjusted the images to a uniform size of $224 \times 224$ to match the network input requirements. To assess performance, datasets were divided into three subsets: training, validation, and test sets. 

\mypar{Unified configuration.} We employ the same hyper-parameter settings for all techniques, as reported in Table \ref{tab:optimizers} and Section \ref{SS:ID}. 

\mypar{Training.} In the training phase, this study adheres to the standard training procedure recommended by \cite{he2016deep}. The use of standard color augmentation aligns with the methodologies outlined in \cite{krizhevsky2012imagenet}. Batch normalization is incorporated immediately following each convolution and prior to activation, as per the guidelines in \cite{ioffe2015batch}. In particular, dropout is not used, consistent with the approach detailed in \cite{ioffe2015batch}. A comprehensive evaluation is performed on 13 models across four families of advanced network frameworks: ConvNeXt, Maxvit, Swin Transformer, and ViT.

\mypar{Evaluation.} In evaluation process, we considered the following performance metrics: Accuracy (ACC), Balanced Accuracy (BACC), Precision (PRE), Recall (REC) and F1 score. Furthermore, the average value of these metrics (AVG) is considered. For calibration, the ECE and reliability plots are used for evaluation.

\section{Experiments}\label{S:4}

\subsection{Datasets}

\mypar{Bladder} The Bladder dataset is derived from a multi-center (C) bladder cancer classification data set using the T2-weighted modality (C1, 160 patients; C2, 48 patients; C3, 32 patients; C4, 35 patients), totaling 279 patients \cite{cao2024multicenter}. It has two classes, namely muscle invasive bladder (MIBC) and non muscle invasive bladder (MNIBC). For this data set, we used data from three centers for training and validation, while the remaining center is used for testing (e.g., ``fold1'' uses C1 to C3 for training (Tr) and validation (Val), and C4 for testing (Te). This results in four partitions, i.e., fold1 (``Tr and Val'': \{C1, C2, C3\}; ``Te'': \{C4\}), fold2 (``Tr and Val'': \{C1, C2, C4\}; ``Te'': \{C3\}), fold3 (``Tr and Val'': \{C1, C3, C4\}; ``Te'': \{C2\}) and fold4 (``Tr and Val'': \{C2, C3, C4\}; ``Te'': \{C1\}). For the training and validation ratio, we set it to 80\%:20\%. Figure \ref{fig:Datasets} depicts the examples (image with ground truth) from each center.

\begin{figure}[!ht]
    \centering
    \includegraphics[width=1\linewidth]{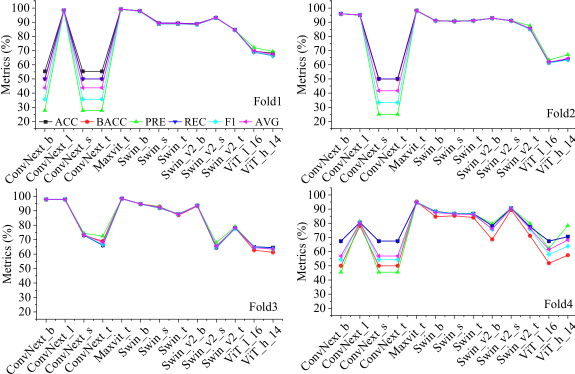}
    \caption{Line chart of validation metrics using 13 deep models on Bladder dataset with SGD optimizer.}
    \label{fig:Bladder_Spider_VAL}
\end{figure}

\begin{figure*}[!ht]
    \centering
    \includegraphics[width=0.195\linewidth]{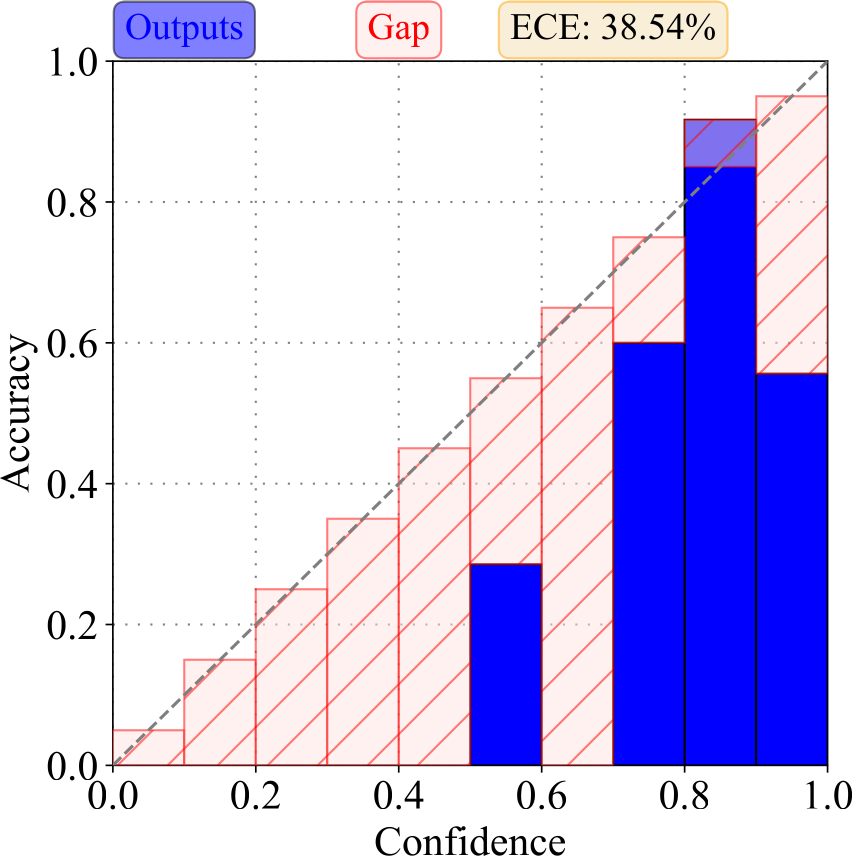} \includegraphics[width=0.195\linewidth]{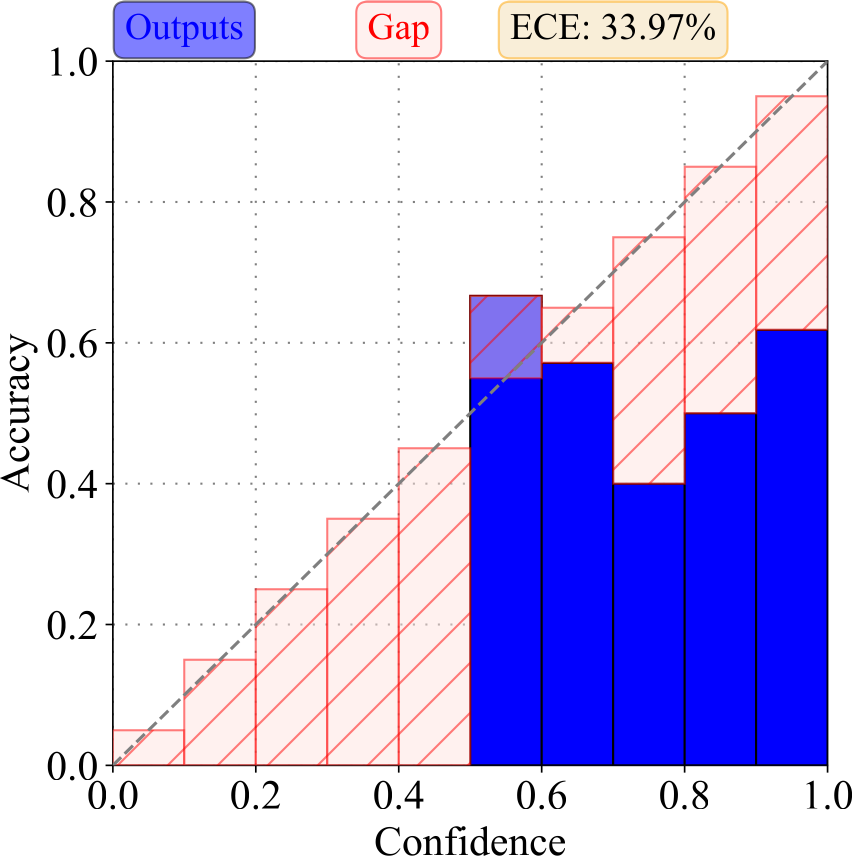} \includegraphics[width=0.195\linewidth]{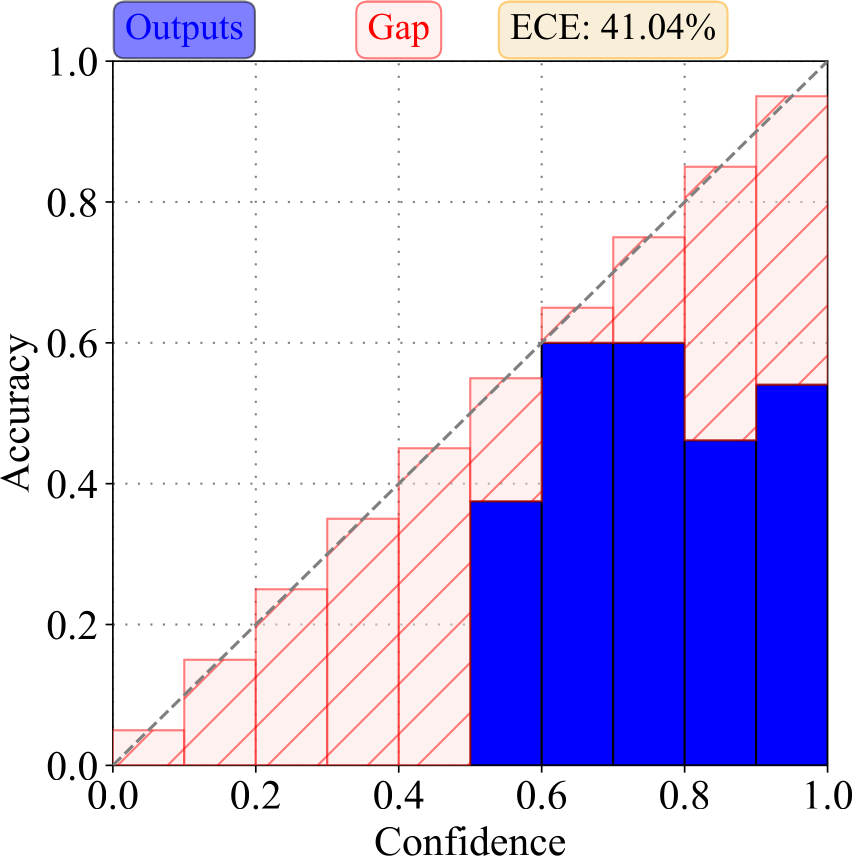} \includegraphics[width=0.195\linewidth]{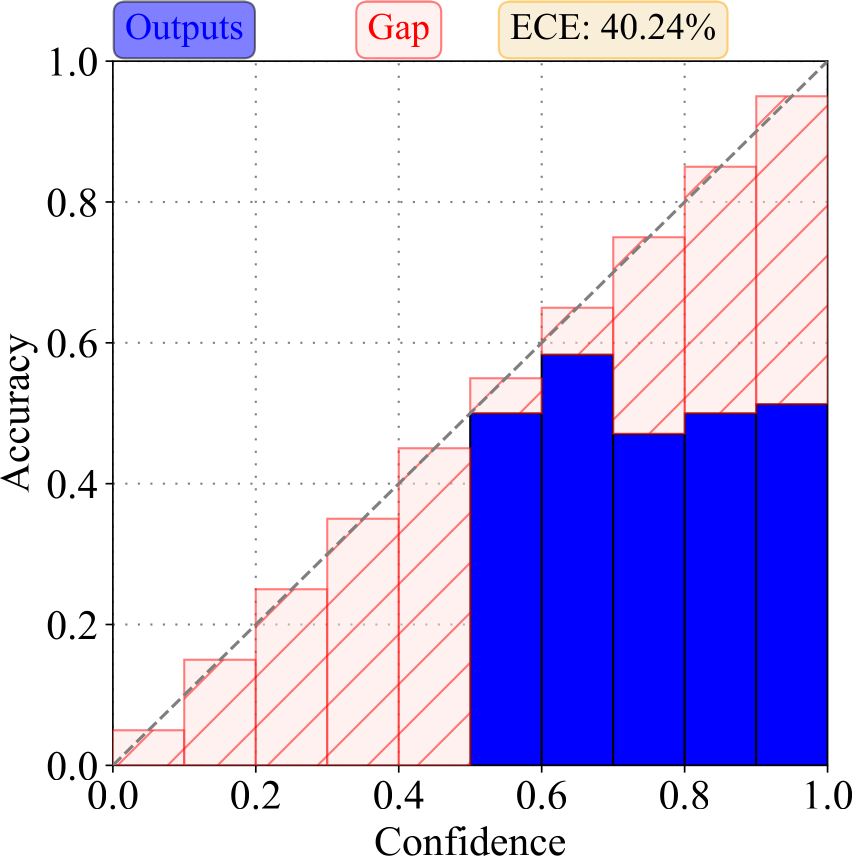} \includegraphics[width=0.195\linewidth]{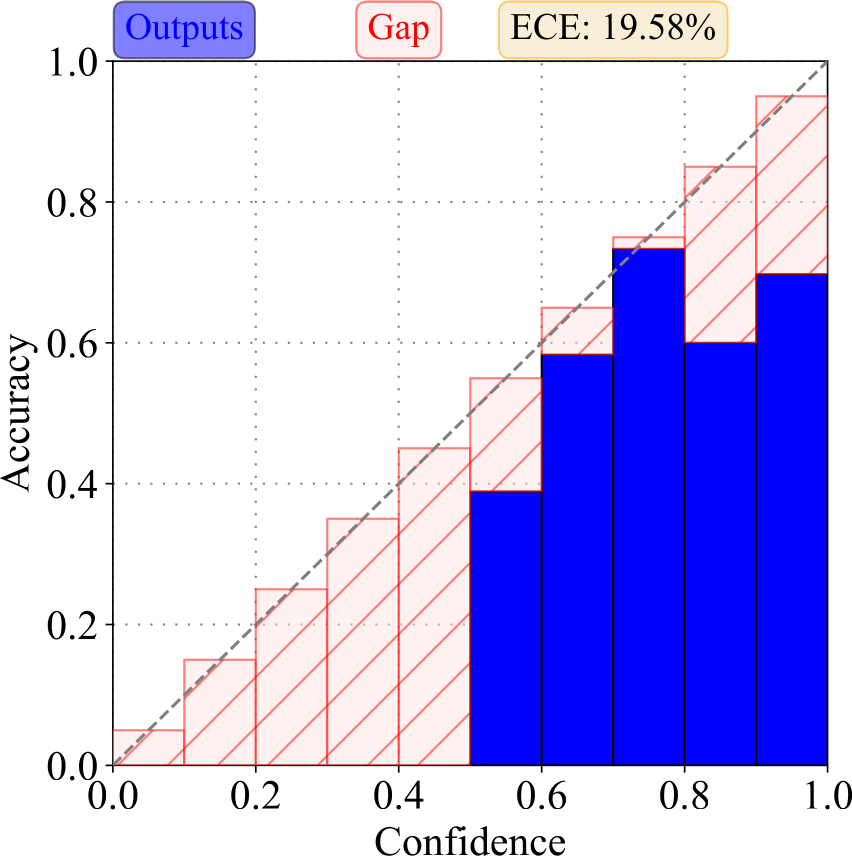}\\ \includegraphics[width=0.195\linewidth]{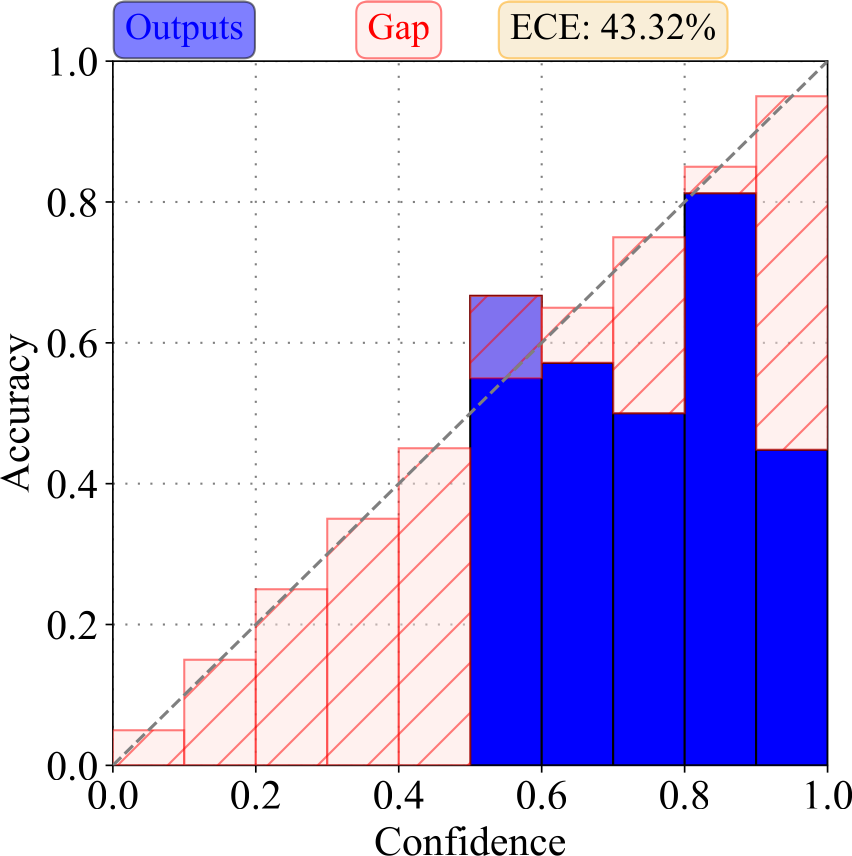} \includegraphics[width=0.195\linewidth]{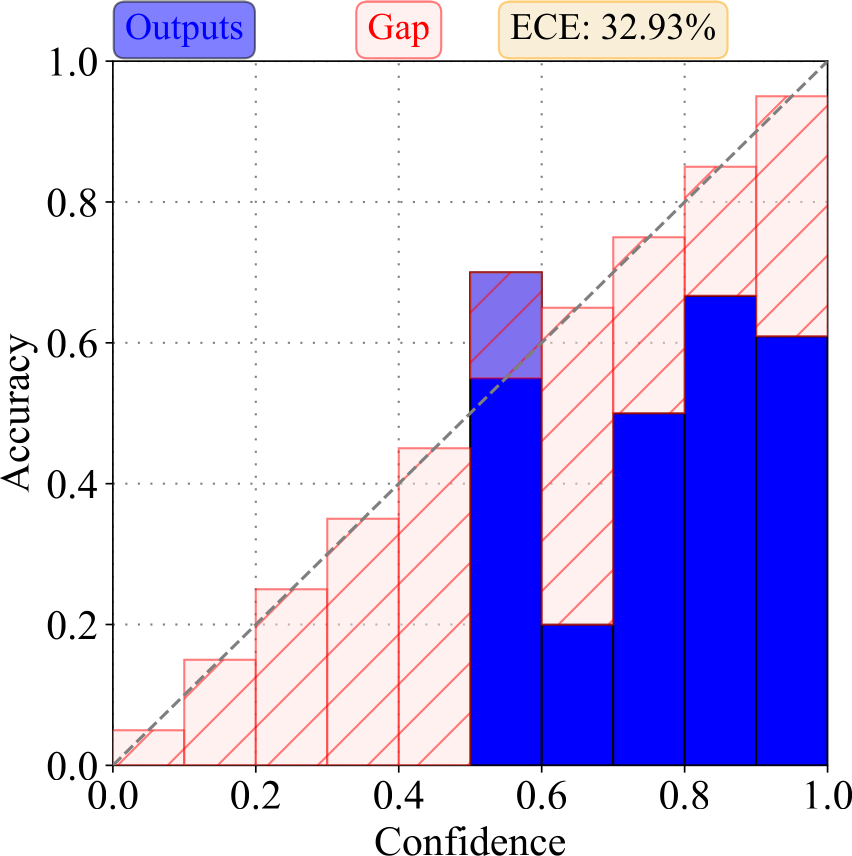} \includegraphics[width=0.195\linewidth]{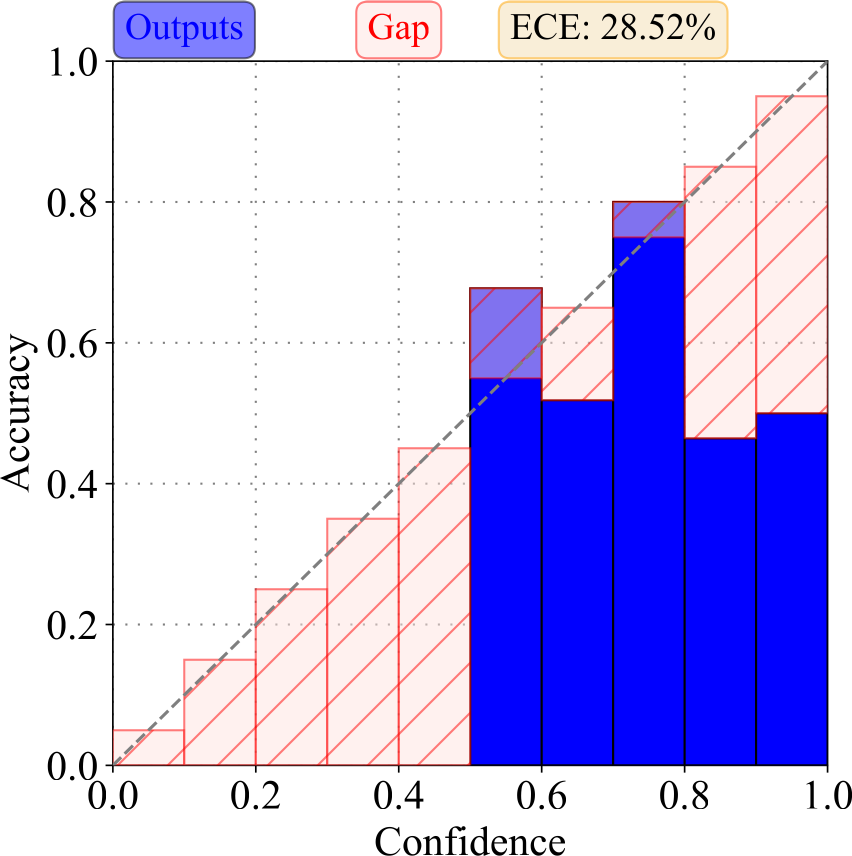} \includegraphics[width=0.195\linewidth]{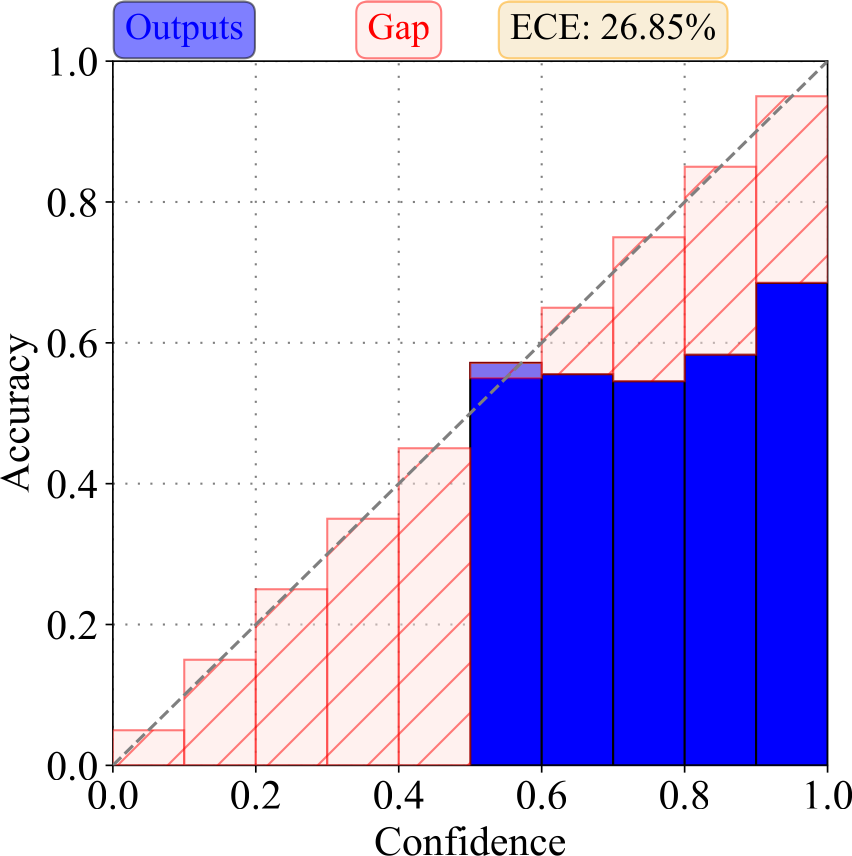} \includegraphics[width=0.195\linewidth]{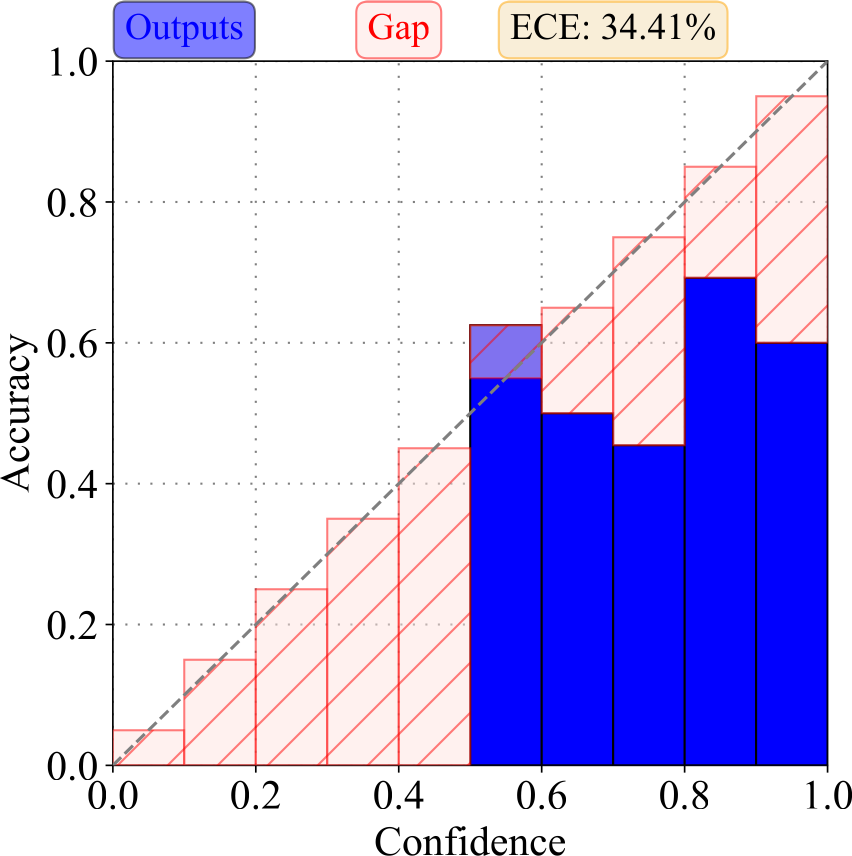}\\
    \includegraphics[width=0.195\linewidth]{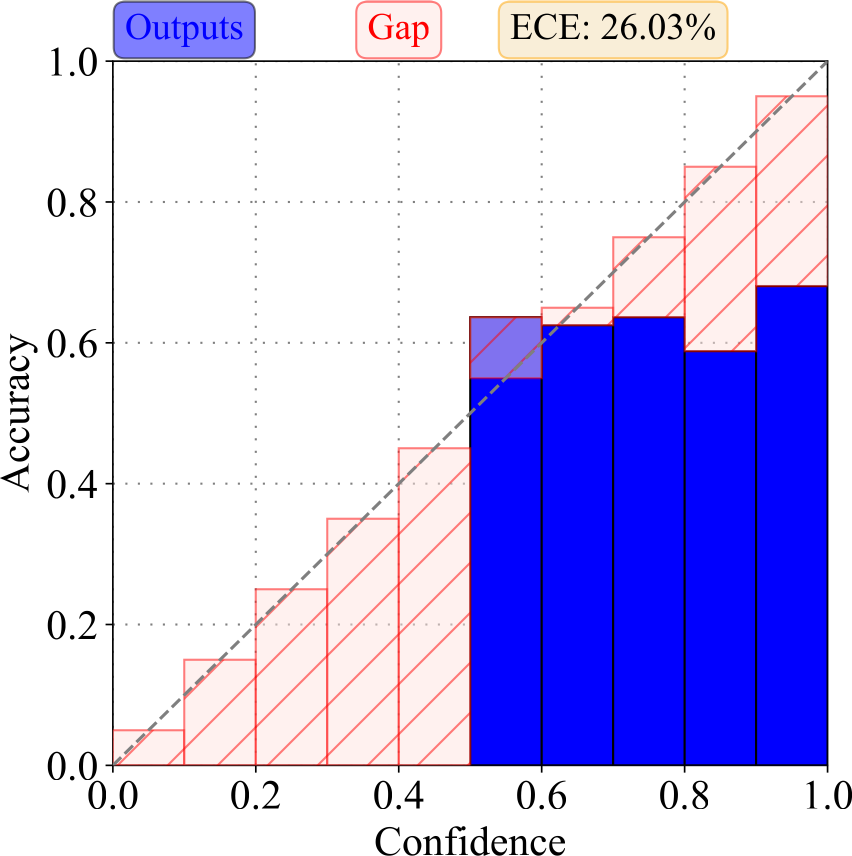} \includegraphics[width=0.195\linewidth]{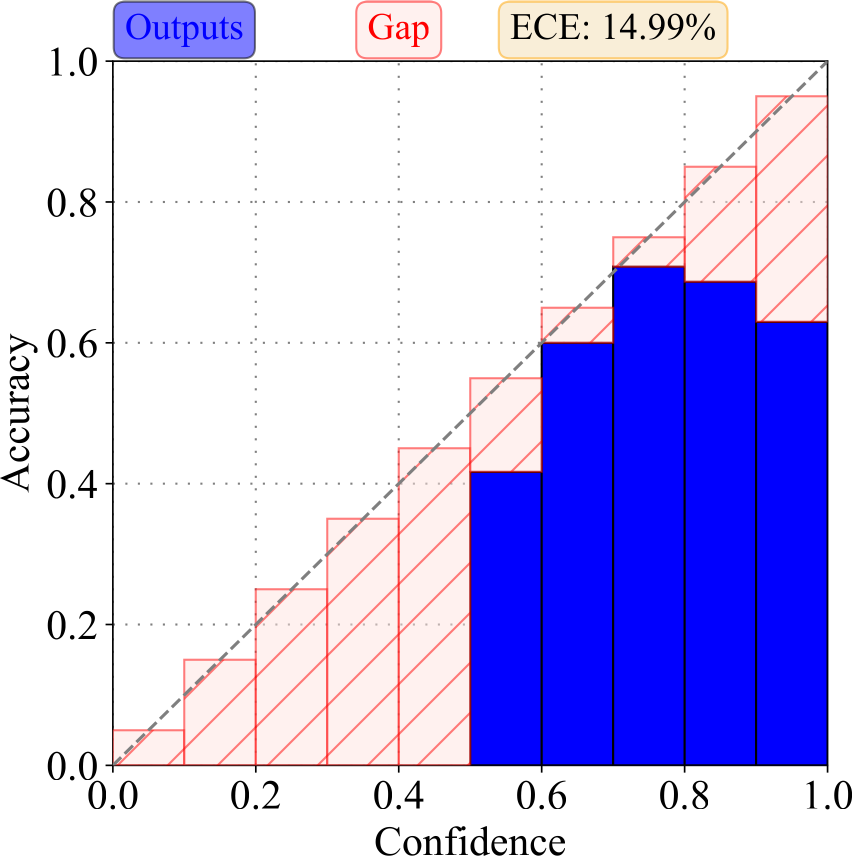} \includegraphics[width=0.195\linewidth]{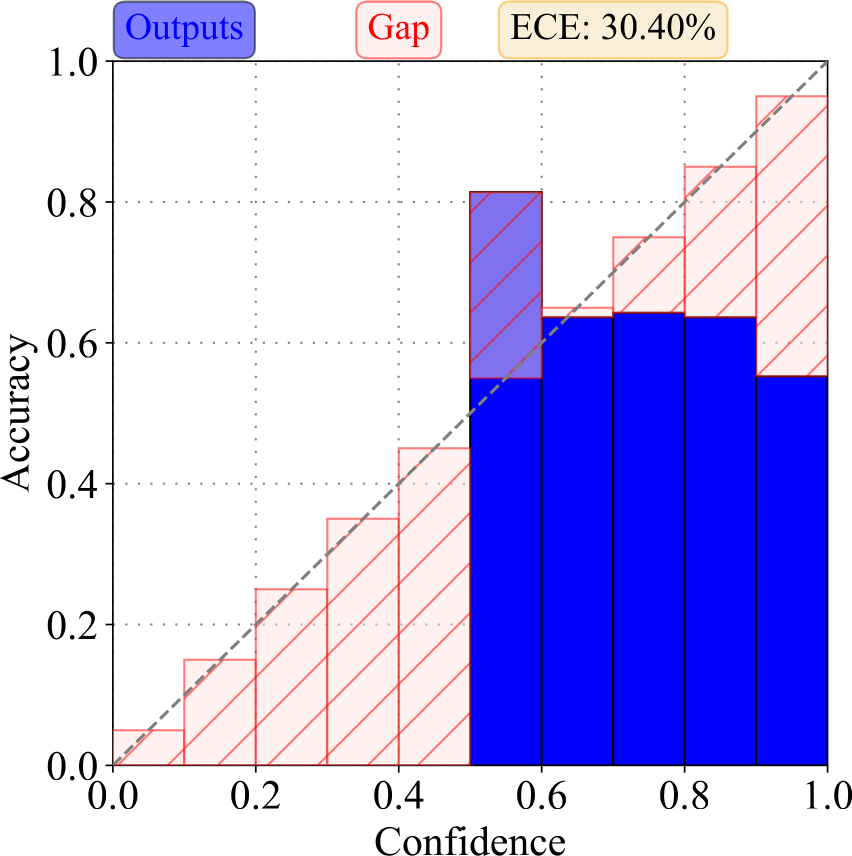}
    \caption{Reliability plot with ECE values for 13 deep models on Bladder dataset using Adagrad optimizer. The first row shows the results of CN\_b, CN\_l, CN\_s, CN\_t and Maxvit\_t, the second row indicates the results of Swin\_b, Swin\_s, Swin\_t, Swin\_v2\_b and Swin\_v2\_s, while the last row represents the results of Swin\_v2\_t, Vit\_l\_16 and Vit\_h\_14.}
    \label{fig:Calibration_Results_Prostate}
\end{figure*}

\subsection{Inplementation details} \label{SS:ID}
We choose the SGD, Adam, AdamW, Adagrad and Adadelta optimizers for simulations \cite{abdulkadirov2023survey}. Table \ref{tab:optimizers} reports a detailed hyper-parameter settings for those optimizers. We considered three architectures, namely convolution based \cite{woo2023convnext} (convnext\_b (CN\_b), convnext\_l (CN\_l), convnext\_s (CN\_s), convnext\_t (CN\_t)), vit based \cite{tu2022maxvit, dosovitskiy2020image} (maxvit\_tiny, vit\_h\_14 and vit\_l\_16), and swin transformer based deep models \cite{liu2021swin} (swin\_b, swin\_s, swin\_t, swin\_v2\_b, swin\_v2\_s, swin\_v2\_t). The cross-entropy function is used to calculate the classification loss. The training epoch is set to 50. The batch size is set to 32, except for vit\_h\_14, which is set to 16. The experiment environment is based on the Windows 11 operating system and features an Intel 13900KF CPU with 128 GB of RAM and an RTX 4090 GPU. We use Pytorch 1.13.1 and Python 3.8. 

\subsection{Tasks}
To provide a comprehensive evaluation of these deep models, we propose three experiments.

\noindent\mypar{Task 1.} We aim to evaluate the models performance using five optimizers. The datasets for this task is Bladder.


\noindent\mypar{Task 2.}
We perform calibration analysis to validate whether these deep models are calibrated without introducing calibration techniques on Bladder dataset.

\begin{figure*}[!ht]
    \centering
    \includegraphics[width=0.8\linewidth]{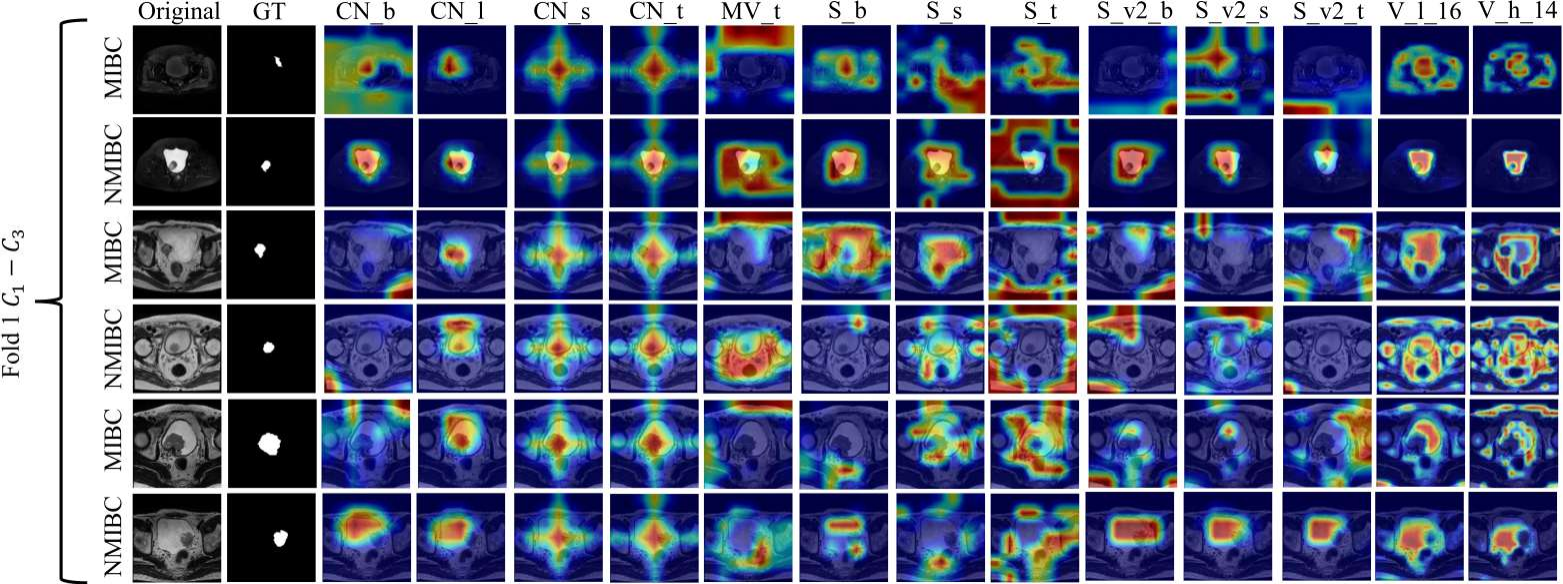}\\
    \caption{Visualizations of the heatmaps for 13 deep models in in-distribution setting. Red indicates that the model focuses more on these regions, while blue indicates a lack of attention. GT denotes the ground truth.}
    \label{fig:XAI_Fold1}
\end{figure*}

\begin{figure*}[!ht]
    \centering
    \includegraphics[width=0.8\linewidth]{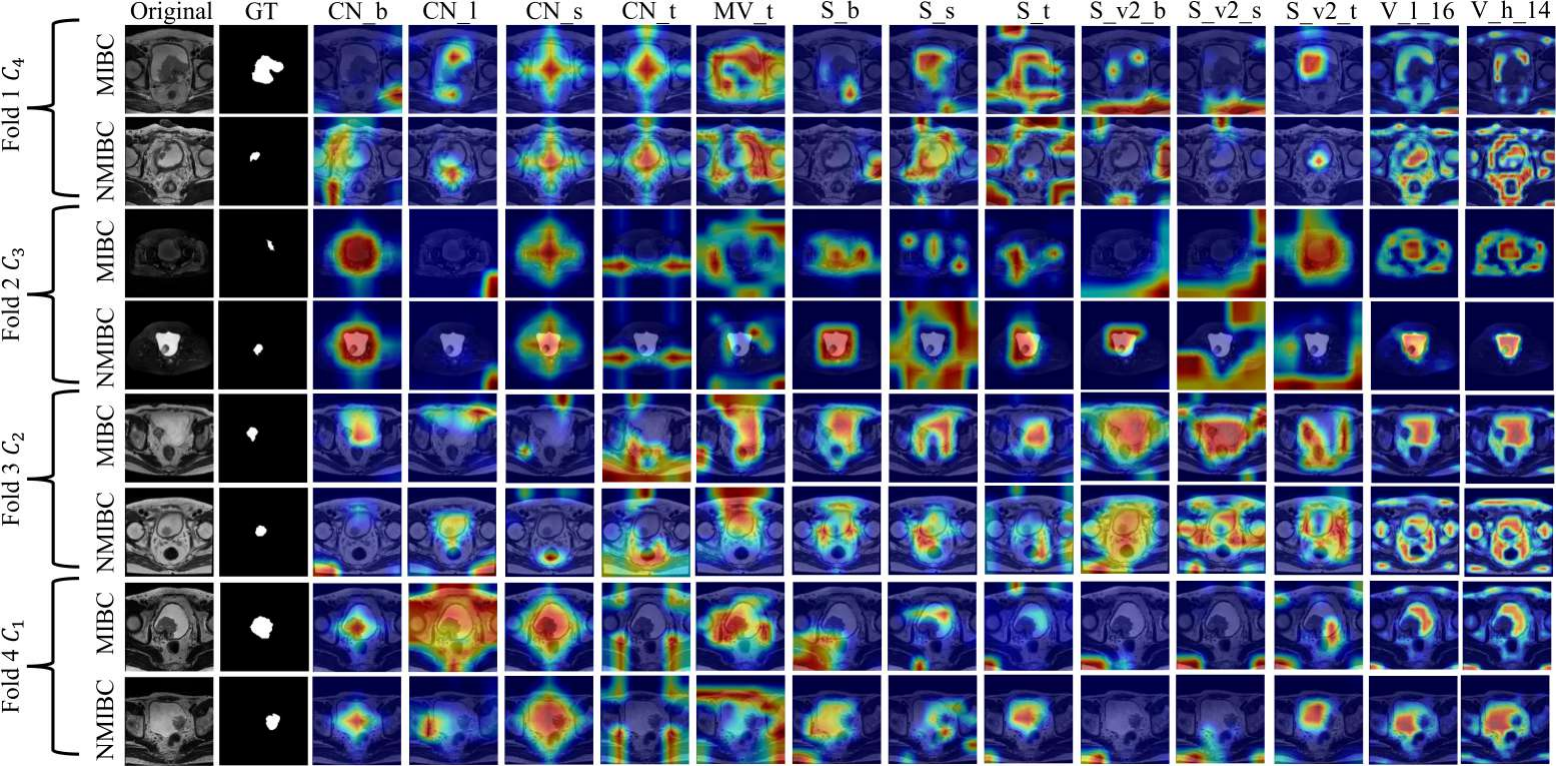}\\
    \caption{Visualizations of the heatmaps for 13 deep models in out-of-distribution setting. GT denotes the ground truth.}
    \label{fig:XAI}
\end{figure*}

\begin{figure*}[!ht]
    \centering
    \includegraphics[width=0.8\linewidth]{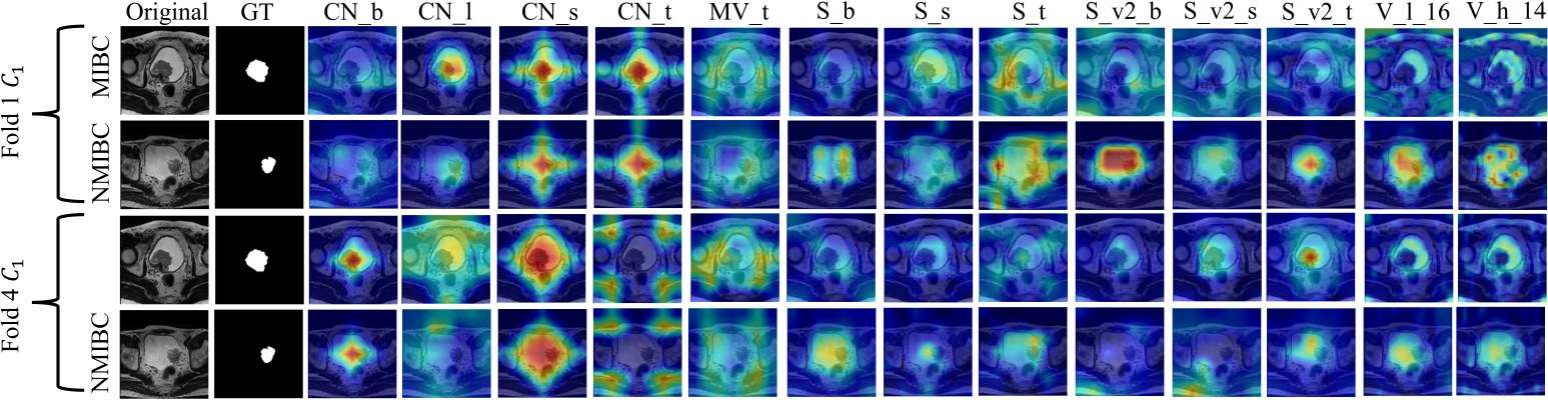}\\
    \caption{Visualizations of the heatmaps for 13 deep models in ID (\textbf{first row}) and OOD (\textbf{second row}) settings after TTA. GT denotes the ground truth.}
    \label{fig:XAI_TTA}
\end{figure*}

\begin{figure}[!ht]
    \centering
    \includegraphics[width=0.475\linewidth]{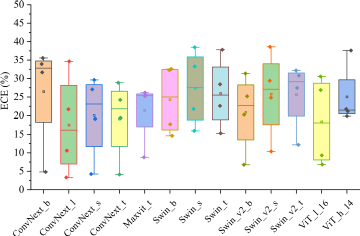} \includegraphics[width=0.475\linewidth]{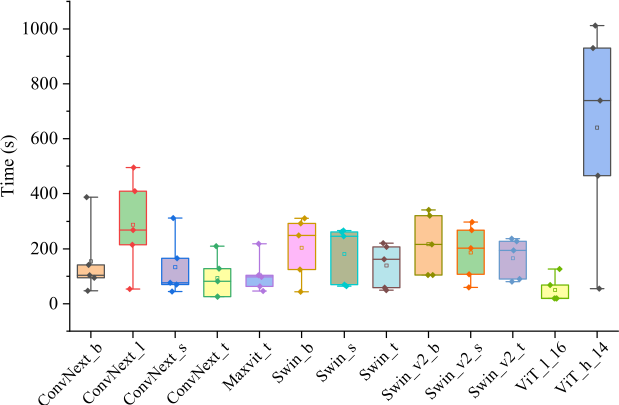}
    \caption{ECE value (\textbf{left}) and execution time (\textbf{right}) for 13 deep models on Bladder using box chart with four (five) optimizers. The diamond represents the ECE value (execution time) of each center (optimizer), while the square indicates the average ECE value (execution time).}
    \label{fig:Calibration_Box}
\end{figure}

\subsection{Task 1 Results}

\noindent \mypar{SGD} For bladder cancer images, as illustrated in Figure \ref{fig:Bladder_Spider_VAL} and Figure \ref{fig:Adagrad_Bladder_Spider}, despite its remarkable validation metrics (e.g., $\sim$ 95\% ACC on Fold2 using CN\_l) using the validation set, its generalization performance on the test set is limited (e.g., only 47.5\% ACC on Fold2 test set). This indicates a gap between in-domain (i.e., same data distribution) and out-of-domain (i.e., different data distribution). Introducing domain adaptation techniques can further improve the performance of these deep models \cite{wu2024facmic}.

\noindent \mypar{Adam} Similar to SGD, all deep models exhibit less than 70\% test ACC on Fold1, except ViT\_h\_14, which has a test ACC of $\sim$ 75\%, as shown in Figure \ref{fig:Adagrad_Bladder_Spider}. However, ViT\_h\_14 and ViT\_l\_16 provide limited test ACC $\sim$ 33\% on Fold4, indicating poor generalization ability. This suggests that using limited training data can lead to severe overfitting for ViTs.

\noindent \mypar{AdamW} As exhibited in Figure \ref{fig:Adagrad_Bladder_Spider}, again, the ViT\_h\_14 and ViT\_l\_16 indicate poor generalization ability on Fold4, consistent with the results of Adam.

\noindent \mypar{Adagrad} The swin transformer based models demonstrate better overall test metrics on Fold1 to Fold4 (e.g., large circle size in the spider plot as illustrated in Figure \ref{fig:Adagrad_Bladder_Spider}).

\noindent \mypar{Adadelta} As shown in Figure \ref{fig:Adagrad_Bladder_Spider}, the use of CN\_b, CN\_l, CN\_t, swin\_b, and swin\_v2\_b show poor generalization ACC (e.g., 32.92\% using CN\_b) on Fold4, limiting their potential where training data is less.

\subsection{Task 2 results}
Classification calibration indicates that the predicted probabilities of a model match the true probability, increasing the reliability of probabilistic predictions \cite{penso2024confidence}. The Bladder ``Fold1'' is used to evaluate these deep models calibration effects. The expected calibration error (ECE) value is measured using only the test set.

Figure \ref{fig:Calibration_Results_Prostate} shows the reliability plot with ECE values for 13 deep models using the Adagrad optimizer on ``Fold1''. As illustrated, despite swin\_v2\_b achieving a higher AVG metric (65.54\%) compared to vit\_l\_16 (62.54\%), the ECE value is much higher compared to vit\_l\_16 (34.41\% vs. 14.99\%). This suggests that swin transformer-based models are less calibrated in bladder cancer classification datasets. In addition, the convnext series all indicate a high ECE value ($\sim$ 40\%).  Furthermore, Figure \ref{fig:Calibration_Box} illustrates the ECE value of the remaining four optimizers using a box plot. Regardless of the optimizer used, Maxvit\_t shows less variance as the boxing length is short on the Bladder dataset. Conversely, CN\_l and CN\_s show the largest box length on the Bladder and ISIC2019 datasets, respectively. This finding highlights the importance of choosing the optimal optimizers.

\subsection{Execution time}
The training time to reach the best validation model is compared to provide a better evaluation of these deep models. The ISIC2019 and Bladder (``Fold1'') datasets are used for evaluation. Figure \ref{fig:Calibration_Box} illustrates the time (s) using a box plot. For example, on Bladder dataset, Maxvit\_t shows the time efficient as its box length is short, while ViT\_h\_14 has the largest box length, $\sim 7\times$ time more than the Maxvit\_t. This result is also consistent with the time measured in the ISIC2019 dataset. Furthermore, Maxvit\_t provides better test metrics using AdamW or SGD compared to ViT\_h\_14 with $\sim 5\times$ less time used, highlighting its potential for skin cancer classification.

\subsection{Interpretability analysis}
The interpretability of model predictions is critical to understanding and trusting deep learning models \cite{chaddad2024generalizable}. It explains how decisions are made to ensure model reliability in clinical applications. We use GradCAM++ \cite{chattopadhay2018grad} to visualize the key features obtained by these models. We validate them on both in-distribution (ID) and out-of-distribution (OOD) settings. Specifically, for ID (e.g., Fold1), we use images derived from $C_1,C_2$ and $C_3$ for experiments, while for OOD (e.g., Fold1) we consider images from $C_4$. 

\mypar{In distribution.} Figure \ref{fig:XAI_Fold1} illustrates the heatmaps obtained from these deep models for the ID setting. For example, all models except Swin\_t show feasible attention to the abnormal regions for the NMIBC class (\textbf{second row}), but for MIBC (\textbf{first row}), Convnext families provide better visual explanations compared to transofmer based models. Similarly, for MIBC derived from $C_1$, Convnext models such as CN\_b provide a reasonable focus on the cancer area, while Maxvit\_t and swin transformer series show scattered attention. Overall, Convnext series show more reliable predictions compared to transformer based for ID condition.

\mypar{Out of distribution.} Figure \ref{fig:XAI} shows the heatmaps obtained from 13 deep models for the OOD condition. For centers that have considerable feature shifts compared to the centers (e.g., $C_3$), most of these models, such as CN\_l, CN\_t, S\_v2\_b, and S\_v2\_s, provide scattered attention across the image for the MIBC and NMIBC classes, suggesting that they are unable to highlight the abnormal regions indicated by the ground truth masks. In contrast, ViT\_l\_16 and ViT\_h\_14 show reasonable focus on abnormal cancer regions, which illuminates the reliability of the predictions. Influenced by various factors such as cancer morphology and textures, we observe that no single model performs optimally under all conditions; it outperforms in certain centers while underperforming in others.

\mypar{Test time augmentation.} We further introduce test time augmentation (TTA) \cite{oza2024breast} as a solution to augment the images during inference to improve the interpretability of these models. Specifically, we consider horizontal flip, rotation, and vertical flip to augment the images. The final heatmap is obtained by taking the average value derived from these augmented images. Figure \ref{fig:XAI_TTA} shows the visualizations of the heatmaps using GradCAM++ after TTA. As illustrated, for ID setting, the use of TTA shifts the attention to the abnormal regions for CN\_b, Maxvit\_t and ViT series for NMIBC. However, for models that pay accurate attention to these regions, the potential of TTA is limited. For OOD, the potential for interpreting NMIBC with CN\_l and CN\_t is limited due to the distribution shifts, but it leads to better visual explanations with Swin\_b, Swin\_v2\_b, and Vit series. These results suggest that TTA is useful for transformer models.

\section{Conclusion}\label{S:6}
This paper presented a comparative simulation of deep models for the classification of images of bladder cancer. It used a multicenter bladder cancer dataset with five optimizers to assess the impact of optimization techniques. Furthermore, we performed calibration analysis on bladder cancer images to evaluate these deep models calibration effects. Finally, we introduced GradCAM++ to evaluate the interpretability of these models, together with TTA to improve the visual explanations. The findings suggest that Adam and AdamW are more suitable for convnext based models, while Adagrad is more convenient for swin transformer models. ViT\_h\_14 also indicates better calibration results compared to the ConvNext series. In addition, these deep models still need domain adaptation \cite{10835760} to improve their interpretability.


{\small
\bibliographystyle{IEEEtran}
\bibliography{ref}

@article{cao2024multicenter,
title={A multicenter bladder cancer MRI dataset and baseline evaluation of federated learning in clinical application},
author={Cao, Kangyang and Zou, Yujian and Zhang, Chang and Zhang, Weijing and Zhang, Jie and Wang, Guojie and Zhang, Chu and Lyu, Jiegeng and Sun, Yue and Zhang, Hongyuan and others},
journal={Scientific Data},
volume={11},
number={1},
pages={1147},
year={2024},
publisher={Nature Publishing Group UK London}
}

@ARTICLE{10835760,
  author={Chaddad, Ahmad and Wu, Yihang and Jiang, Yuchen and Bouridane, Ahmed and Desrosiers, Christian},
  journal={IEEE Transactions on Instrumentation and Measurement}, 
  title={Simulations of Common Unsupervised Domain Adaptation Algorithms for Image Classification}, 
  year={2025},
  volume={74},
  number={},
  pages={1-17},
  doi={10.1109/TIM.2025.3527531}}

@ARTICLE{10902405,
  author={Wu, Yihang and Chaddad, Ahmad and Desrosiers, Christian and Daqqaq, Tareef and Kateb, Reem},
  journal={IEEE Internet of Things Journal}, 
  title={FAA-CLIP: Federated Adversarial Adaptation of CLIP}, 
  year={2025},
  volume={12},
  number={12},
  pages={21091-21102},
  doi={10.1109/JIOT.2025.3545574}}

@article{el2025accurate,
  title={Accurate bladder cancer diagnosis using ensemble deep leaning},
  author={El-Atier, Rana A and Saraya, MS and Saleh, Ahmed I and Rabie, Asmaa H},
  journal={Scientific Reports},
  volume={15},
  number={1},
  pages={12880},
  year={2025},
  publisher={Nature Publishing Group UK London}
}

@article{peng2025stacking,
  title={A stacking ensemble system for identifying the presence of histological variants in bladder carcinoma: a multicenter study},
  author={Peng, Canjie and He, Quanhao and Lv, Fajin and Jiang, Qing and Chen, Yong and Wei, Zongjie and Xv, Yingjie and Liao, Fangtong and Xiao, Mingzhao},
  journal={Frontiers in Oncology},
  volume={14},
  pages={1469427},
  year={2025}
}

@article{zou2025prediction,
  title={Prediction of muscular-invasive bladder cancer using multi-view fusion self-distillation model based on 3D T2-Weighted images},
  author={Zou, Yuan and Yu, Jie and Cai, Lingkai and Chen, Chunxiao and Meng, Ruoyu and Xiao, Yueyue and Fu, Xue and Yang, Xiao and Liu, Peikun and Lu, Qiang},
  journal={Biomedical Engineering/Biomedizinische Technik},
  volume={70},
  number={1},
  pages={37--47},
  year={2025},
  publisher={De Gruyter}
}

@inproceedings{chattopadhay2018grad,
  title={Grad-cam++: Generalized gradient-based visual explanations for deep convolutional networks},
  author={Chattopadhay, Aditya and Sarkar, Anirban and Howlader, Prantik and Balasubramanian, Vineeth N},
  booktitle={2018 IEEE winter conference on applications of computer vision (WACV)},
  pages={839--847},
  year={2018},
  organization={IEEE}
}

@article{chaddad2024generalizable,
  title={Generalizable and explainable deep learning for medical image computing: An overview},
  author={Chaddad, Ahmad and Hu, Yan and Wu, Yihang and Wen, Binbin and Kateb, Reem},
  journal={Current Opinion in Biomedical Engineering},
  pages={100567},
  year={2024},
  publisher={Elsevier}
}

@article{oza2024breast,
  title={Breast lesion classification from mammograms using deep neural network and test-time augmentation},
  author={Oza, Parita and Sharma, Paawan and Patel, Samir},
  journal={Neural Computing and Applications},
  volume={36},
  number={4},
  pages={2101--2117},
  year={2024},
  publisher={Springer}
}

@article{zhang2025texture,
  title={Texture graph transformer for prostate cancer classification},
  author={Zhang, Guokai and Gao, Lin and Liu, Huan and Wang, Shuihua and Xu, Xiaowen and Zhao, Binghui},
  journal={Biomedical Signal Processing and Control},
  volume={99},
  pages={106890},
  year={2025},
  publisher={Elsevier}
}

@article{li2023predicting,
  title={Predicting muscle invasion in bladder cancer based on MRI: A comparison of radiomics, and single-task and multi-task deep learning},
  author={Li, Jianpeng and Qiu, Zhengxuan and Cao, Kangyang and Deng, Lei and Zhang, Weijing and Xie, Chuanmiao and Yang, Shuiqing and Yue, Peiyan and Zhong, Jian and Lyu, Jiegeng and others},
  journal={Computer Methods and Programs in Biomedicine},
  volume={233},
  pages={107466},
  year={2023},
  publisher={Elsevier}
}

@article{khan2023skinvit,
  title={SkinViT: A transformer based method for Melanoma and Nonmelanoma classification},
  author={Khan, Somaiya and Khan, Ali},
  journal={Plos one},
  volume={18},
  number={12},
  pages={e0295151},
  year={2023},
  publisher={Public Library of Science San Francisco, CA USA}
}

@inproceedings{rippa2024classification,
  title={Classification of Prostate Cancer in 3D Magnetic Resonance Imaging Data based on Convolutional Neural Networks},
  author={Rippa, Malte and Schulze, Ruben and Himstedt, Marian and Burn, Felice},
  booktitle={Current Directions in Biomedical Engineering},
  volume={10},
  number={1},
  pages={61--64},
  year={2024},
  organization={De Gruyter}
}

@article{abdulkadirov2023survey,
  title={Survey of optimization algorithms in modern neural networks},
  author={Abdulkadirov, Ruslan and Lyakhov, Pavel and Nagornov, Nikolay},
  journal={Mathematics},
  volume={11},
  number={11},
  pages={2466},
  year={2023},
  publisher={MDPI}
}

@inproceedings{wu2024facmic,
  title={Facmic: Federated adaptative clip model for medical image classification},
  author={Wu, Yihang and Desrosiers, Christian and Chaddad, Ahmad},
  booktitle={International Conference on Medical Image Computing and Computer-Assisted Intervention},
  pages={531--541},
  year={2024},
  organization={Springer}
}

@article{penso2024confidence,
  title={Confidence calibration of a medical imaging classification system that is robust to label noise},
  author={Penso, Coby and Frenkel, Lior and Goldberger, Jacob},
  journal={IEEE Transactions on Medical Imaging},
  year={2024},
  publisher={IEEE}
}

@inproceedings{wang2024sscd,
  title={SSCD-Net: Semi-supervised Skin Cancer Diagnostical Network Combined with Curriculum Learning, Disease Relation and Clinical Information},
  author={Wang, Wei and Cao, CaoYunjian and Wu, Shaozhi and Liu, Xingang and Su, Han and Tian, Dan},
  booktitle={2024 International Joint Conference on Neural Networks (IJCNN)},
  pages={1--8},
  year={2024},
  organization={IEEE}
}

@inproceedings{woo2023convnext,
  title={Convnext v2: Co-designing and scaling convnets with masked autoencoders},
  author={Woo, Sanghyun and Debnath, Shoubhik and Hu, Ronghang and Chen, Xinlei and Liu, Zhuang and Kweon, In So and Xie, Saining},
  booktitle={Proceedings of the IEEE/CVF Conference on Computer Vision and Pattern Recognition},
  pages={16133--16142},
  year={2023}
}

@inproceedings{tu2022maxvit,
  title={Maxvit: Multi-axis vision transformer},
  author={Tu, Zhengzhong and Talebi, Hossein and Zhang, Han and Yang, Feng and Milanfar, Peyman and Bovik, Alan and Li, Yinxiao},
  booktitle={European conference on computer vision},
  pages={459--479},
  year={2022},
  organization={Springer}
}

@inproceedings{liu2021swin,
  title={Swin transformer: Hierarchical vision transformer using shifted windows},
  author={Liu, Ze and Lin, Yutong and Cao, Yue and Hu, Han and Wei, Yixuan and Zhang, Zheng and Lin, Stephen and Guo, Baining},
  booktitle={Proceedings of the IEEE/CVF international conference on computer vision},
  pages={10012--10022},
  year={2021}
}

@article{he2016deep,
  title={Deep residual learning for image recognition},
  author={He, Kaiming and Zhang, Xiangyu and Ren, Shaoqing and Sun, Jian},
  journal={Proceedings of the IEEE conference on computer vision and pattern recognition},
  pages={770--778},
  year={2016}
}

@article{chaddad2023federated,
  title={Federated learning for healthcare applications},
  author={Chaddad, Ahmad and Wu, Yihang and Desrosiers, Christian},
  journal={IEEE Internet of Things Journal},
  year={2023},
  publisher={IEEE}
}

@article{dosovitskiy2020image,
  title={An image is worth 16x16 words: Transformers for image recognition at scale},
  author={Dosovitskiy, Alexey and Beyer, Lucas and Kolesnikov, Alexander and Weissenborn, Dirk and Zhai, Xiaohua and Unterthiner, Thomas and Dehghani, Mostafa and Minderer, Matthias and Heigold, Georg and Gelly, Sylvain and others},
  journal={arXiv preprint arXiv:2010.11929},
  year={2020}
}

@article{bashkami2024review,
  title={A review of Artificial Intelligence methods in bladder cancer: segmentation, classification, and detection},
  author={Bashkami, Ayah and Nasayreh, Ahmad and Makhadmeh, Sharif Naser and Gharaibeh, Hasan and Alzahrani, Ahmed Ibrahim and Alwadain, Ayed and Heming, Jia and Ezugwu, Absalom E and Abualigah, Laith},
  journal={Artificial Intelligence Review},
  volume={57},
  number={12},
  pages={339},
  year={2024},
  publisher={Springer}
}

@article{liu2021multiscale,
  title={Multiscale ensemble of convolutional neural networks for skin lesion classification},
  author={Liu, Yi-Peng and Wang, Ziming and Li, Zhanqing and Li, Jing and Li, Ting and Chen, Peng and Liang, Ronghua},
  journal={IET Image Processing},
  volume={15},
  number={10},
  pages={2309--2318},
  year={2021},
  publisher={Wiley Online Library}
}

@article{yang2021application,
  title={Application of deep learning as a noninvasive tool to differentiate muscle-invasive bladder cancer and non--muscle-invasive bladder cancer with CT},
  author={Yang, Yuhan and Zou, Xiuhe and Wang, Yixi and Ma, Xuelei},
  journal={European Journal of Radiology},
  volume={139},
  pages={109666},
  year={2021},
  publisher={Elsevier}
}

@article{shalata2024precise,
  title={Precise grading of non-muscle invasive bladder cancer with multi-scale pyramidal CNN},
  author={Shalata, Aya T and Alksas, Ahmed and Shehata, Mohamed and Khater, Sherry and Ezzat, Osama and Ali, Khadiga M and Gondim, Dibson and Mahmoud, Ali and El-Gendy, Eman M and Mohamed, Mohamed A and others},
  journal={Scientific Reports},
  volume={14},
  number={1},
  pages={25131},
  year={2024},
  publisher={Nature Publishing Group UK London}
}

@article{khedr2023classification,
  title={The classification of the bladder cancer based on Vision Transformers (ViT)},
  author={Khedr, Ola S and Wahed, Mohamed E and Al-Attar, Al-Sayed R and Abdel-Rehim, EA},
  journal={Scientific Reports},
  volume={13},
  number={1},
  pages={20639},
  year={2023},
  publisher={Nature Publishing Group UK London}
}

@article{kurata2024development,
  title={Development of deep learning model for diagnosing muscle-invasive bladder cancer on MRI with vision transformer},
  author={Kurata, Yasuhisa and Nishio, Mizuho and Moribata, Yusaku and Otani, Satoshi and Himoto, Yuki and Takahashi, Satoru and Kusakabe, Jiro and Okura, Ryota and Shimizu, Marina and Hidaka, Keisuke and others},
  journal={Heliyon},
  volume={10},
  number={16},
  year={2024},
  publisher={Elsevier}
}

@inproceedings{guergueb2022skin,
  title={Skin cancer detection using ensemble learning and grouping of deep models},
  author={Guergueb, Takfarines and Akhloufi, Moulay A},
  booktitle={Proceedings of the 19th International Conference on Content-based Multimedia Indexing},
  pages={121--125},
  year={2022}
}

@article{jiao2024prediction,
  title={Prediction of HER2 Status Based on Deep Learning in H\&E-Stained Histopathology Images of Bladder Cancer},
  author={Jiao, Panpan and Zheng, Qingyuan and Yang, Rui and Ni, Xinmiao and Wu, Jiejun and Chen, Zhiyuan and Liu, Xiuheng},
  journal={Biomedicines},
  volume={12},
  number={7},
  pages={1583},
  year={2024},
  publisher={MDPI}
}

@article{eminaga2023efficient,
  title={Efficient Augmented Intelligence Framework for Bladder Lesion Detection},
  author={Eminaga, Okyaz and Lee, Timothy Jiyong and Laurie, Mark and Ge, T Jessie and La, Vinh and Long, Jin and Semjonow, Axel and Bogemann, Martin and Lau, Hubert and Shkolyar, Eugene and others},
  journal={JCO Clinical Cancer Informatics},
  volume={7},
  pages={e2300031},
  year={2023},
  publisher={Wolters Kluwer Health}
}

@article{krizhevsky2012imagenet,
  title={Imagenet classification with deep convolutional neural networks},
  author={Krizhevsky, Alex and Sutskever, Ilya and Hinton, Geoffrey E},
  journal={Advances in neural information processing systems},
  volume={25},
  year={2012}
}

@inproceedings{ioffe2015batch,
  title={Batch normalization: Accelerating deep network training by reducing internal covariate shift},
  author={Ioffe, Sergey and Szegedy, Christian},
  booktitle={International conference on machine learning},
  pages={448--456},
  year={2015},
  organization={pmlr}
}
}

\end{document}